%% file: main.tex
\documentclass[conference]{IEEEtran}
\input{preamble}

\begin{document}

\title{Efficient Hierarchical Any-Angle Path Planning\\on Multi-Resolution 3D Grids}

\author{\authorblockN{Victor~Reijgwart, Cesar~Cadena, Roland~Siegwart and Lionel~Ott}
        \authorblockA{Autonomous Systems Lab, ETH Z\"urich, Switzerland\\
        Email: vreijgwart@rai-inst.com, $[$cesarc $\vert$ rolandsi $\vert$ lioott$]$@ethz.ch}}


%

\maketitle

\input{sections/abstract}

\IEEEpeerreviewmaketitle

\setlength{\textfloatsep}{0.7em}
\input{sections/introduction}
\input{sections/related_work}
\input{sections/problem_statement.tex}
\input{sections/method}
\input{sections/experiments}
\input{sections/limitations}
\input{sections/conclusion}
\input{sections/acknowledgements}

\bibliographystyle{plainnat}
\bibliography{references}

\appendix
\input{sections/appendix}

\end{document}

%% file: preamble.tex
\usepackage{times}
\usepackage[numbers]{natbib} 
\usepackage{multicol}
\usepackage[bookmarks=true]{hyperref}

\usepackage[nolist,nohyperlinks]{acronym}
\usepackage{amsmath,amsfonts}
\usepackage{graphicx}
\usepackage{xcolor}
\usepackage{siunitx} 
\usepackage{calc} 
\usepackage[ruled,linesnumbered]{algorithm2e}

\input{acronyms}
\input{commands}

\usepackage[capitalize]{cleveref}
\crefname{line}{Li.}{Lis.}
\Crefname{line}{Line}{Lines}
\crefname{algorithm}{Alg.}{Algs.}
\Crefname{algorithm}{Algorithm}{Algorithms}
\usepackage{multirow}

%% file: acronyms.tex
\begin{acronym}
    \acro{SLAM}{Simultaneous Localization And Mapping}
    \acro{TSDF}{Truncated Signed Distance Field}
    \acro{ESDF}{Euclidean Signed Distance Field}
    \acro{MAV}{Micro Aerial Vehicle}
    \acro{AABB}{Axis Aligned Bounding Box}
    \acrodefplural{AABB}{Axis-Aligned Bounding Boxes}
    \acro{MRA}{Multi-Resolution Analysis}
    \acro{FWT}{Fast Wavelet Transform}
    \acro{ROC}{Receiver Operating Characteristic}
    \acro{AUC}{Area under the ROC Curve}
    \acro{TPR}{True Positive Rate}
    \acro{FPR}{False Positive Rate}
    \acro{RMP}{Riemannian Motion Policy}
    \acrodefplural{RMP}{Riemannian Motion Policies}
    \acro{EDF}{Euclidean Distance Field}
\end{acronym}

%% file: commands.tex
\newcommand{\OpenQueue}{\texttt{open}}
\newcommand{\ClosedSet}{\texttt{closed}}
\newcommand{\Predecessor}{\texttt{predecessor}}
\newcommand{\UpdateSubvolume}{\texttt{UpdateSubvolume}} 
\newcommand{\UpdateVertex}{\texttt{UpdateVertex}} 
\newcommand{\UpdateCost}{\texttt{UpdateCost}}
\newcommand{\ComputeFScore}{\texttt{ComputeFScore}}
\newcommand{\IsBetterOrSimilar}{\texttt{IsBetterOrSimilar}}
\newcommand{\PathFound}{\textit{PathFound}}
\newcommand{\NoPathFound}{\textit{NoPathFound}}
\newcommand{\Changed}{\textit{Changed}}
\newcommand{\Unchanged}{\textit{Unchanged}}
\newcommand{\StrictlyBetter}{\textit{StrictlyBetter}}
\newcommand{\NotBetter}{\textit{NotBetter}}
\newcommand{\Ambiguous}{\textit{Ambiguous}}
\newcommand{\Mine}{\texttt{Mine}}
\newcommand{\Cloister}{\texttt{Cloister}}
\newcommand{\Math}{\texttt{Math}}
\newcommand{\Park}{\texttt{Park}}
\newcommand{\MatchMap}{\texttt{Match\;map}}
\newcommand{\NoExplicitInit}{\texttt{No\:init}}
\newcommand{\Ours}{\textit{Ours}}
\newcommand{\OursLazy}{\textit{OursLazy}}
\newcommand{\OursFast}{\textit{OursFast}}
\newcommand{\AStar}{\texttt{A*}}
\newcommand{\ThetaStar}{\texttt{Theta*}}
\newcommand{\LazyThetaStar}{\texttt{LazyTheta*}}
\newcommand{\OctreeLazyThetaStar}{\texttt{OctreeLazyTheta*}}
\newcommand{\RRT}[1]{\texttt{RRT*\,#1s}}
\newcommand{\RRTConnect}{\texttt{RRTConnect}}

%% file: sections/abstract.tex
\begin{abstract}
Hierarchical, multi-resolution volumetric mapping approaches are widely used to represent large and complex environments as they can efficiently capture their occupancy and connectivity information. Yet widely used path planning methods such as sampling and trajectory optimization do not exploit this explicit connectivity information, and search-based methods such as A* suffer from scalability issues in large-scale high-resolution maps. In many applications, Euclidean shortest paths form the underpinning of the navigation system. For such applications, any-angle planning methods, which find optimal paths by connecting corners of obstacles with straight-line segments, provide a simple and efficient solution. In this paper, we present a method that has the optimality and completeness properties of any-angle planners while overcoming computational tractability issues common to search-based methods by exploiting multi-resolution representations. Extensive experiments on real and synthetic environments demonstrate the proposed approach's solution quality and speed, outperforming even sampling-based methods. The framework is open-sourced to allow the robotics and planning community to build on our research.

\end{abstract}

%% file: sections/introduction.tex
\section{Introduction}
A core competency of robots is the ability to autonomously navigate between areas of interest, such as storage spaces, work sites, and inspection points, even if these locations are far apart. Methods for solving this planning problem can be categorized into optimization-, sampling- and search-based approaches. Optimization-based methods produce high-quality, continuous solutions but generally require an initial guess to converge to a good solution. This initial guess is often obtained from a sampling- or search-based planner. Search-based methods generally operate on a graph with a fixed topology, such as a grid map's adjacency graph or a state lattice constructed from motion primitives. Meanwhile, sampling-based methods build the graph by randomly sampling and connecting collision-free robot configurations. Sampling-based approaches are popular in practice due to their ability to find solutions while only sparsely covering large, potentially high-dimensional configuration spaces. However, extracting graphs through random sampling discards much of the information embedded in the volumetric map and neglects its underlying structure. The information contained in discretized maps is finite, yet sampling-based methods are only asymptotically complete and cannot detect infeasibility in finite time. This is particularly problematic in environments with narrow passages, where solving a planning query can take a long time, and feasibility is not guaranteed. This raises a hard-to-answer question in sampling-based methods: How long should one try to find a solution before giving up?

\begin{figure}[bt]
    \vspace{+0.6em}
    \centering
    \includegraphics[width=\linewidth]{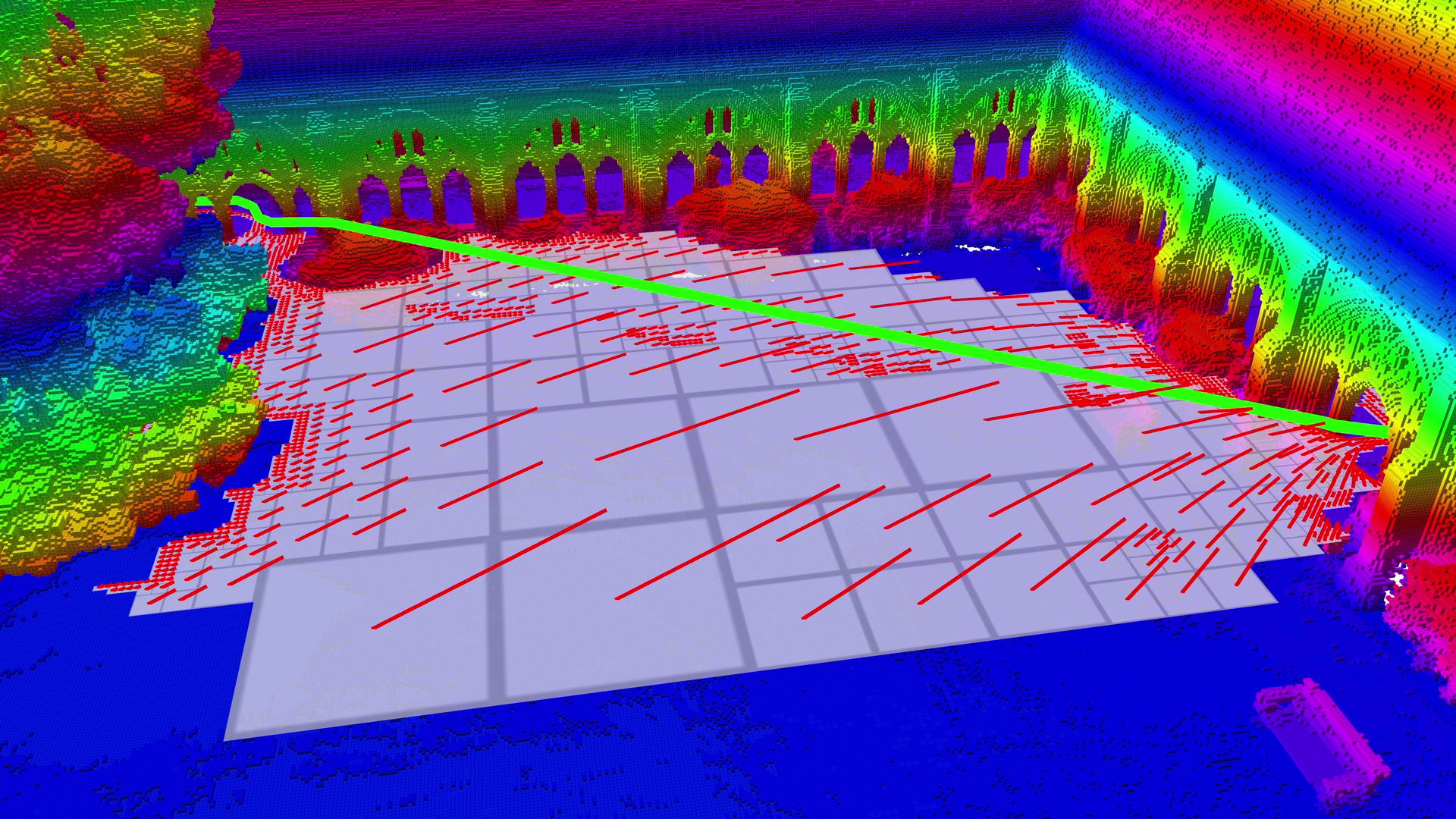}
    \caption{Illustration of a solution generated by our proposed any-angle path planner, \textit{wavestar}. The shortest path ({\color[HTML]{26ff07}green}) consists of straight-line segments that efficiently traverse free space using a small number of waypoints which tightly fit the obstacles (voxels shaded by height). A 2D slice of the multi-resolution 3D cost field demonstrates how the hierarchical algorithm refines resolution only where necessary, ensuring both efficiency and accuracy.}
    \label{fig:global_planning/media/hero_shot}
    \vspace{-0.3em}
\end{figure}

For many applications, a volumetric map's adjacency graph provides a reasonable discretization of the true, continuous search space. Combining this graph with standard search algorithms~\cite{dijkstra1959note} allows resolution-complete solutions to be found in finite time. While searching for the shortest path, A*~\cite{hart1968Astar} and similar methods compute the optimal cost-to-come and predecessor for each explored grid vertex. Their time and space complexity, therefore, scales linearly with the explored volume and cubically with the grid resolution~\cite{yatziv2006oNFastMarching}. This is particularly problematic in environments with dead-ends that are deep relative to the grid resolution.

\begin{figure}[bt]
    \centering
    \includegraphics[width=\linewidth]{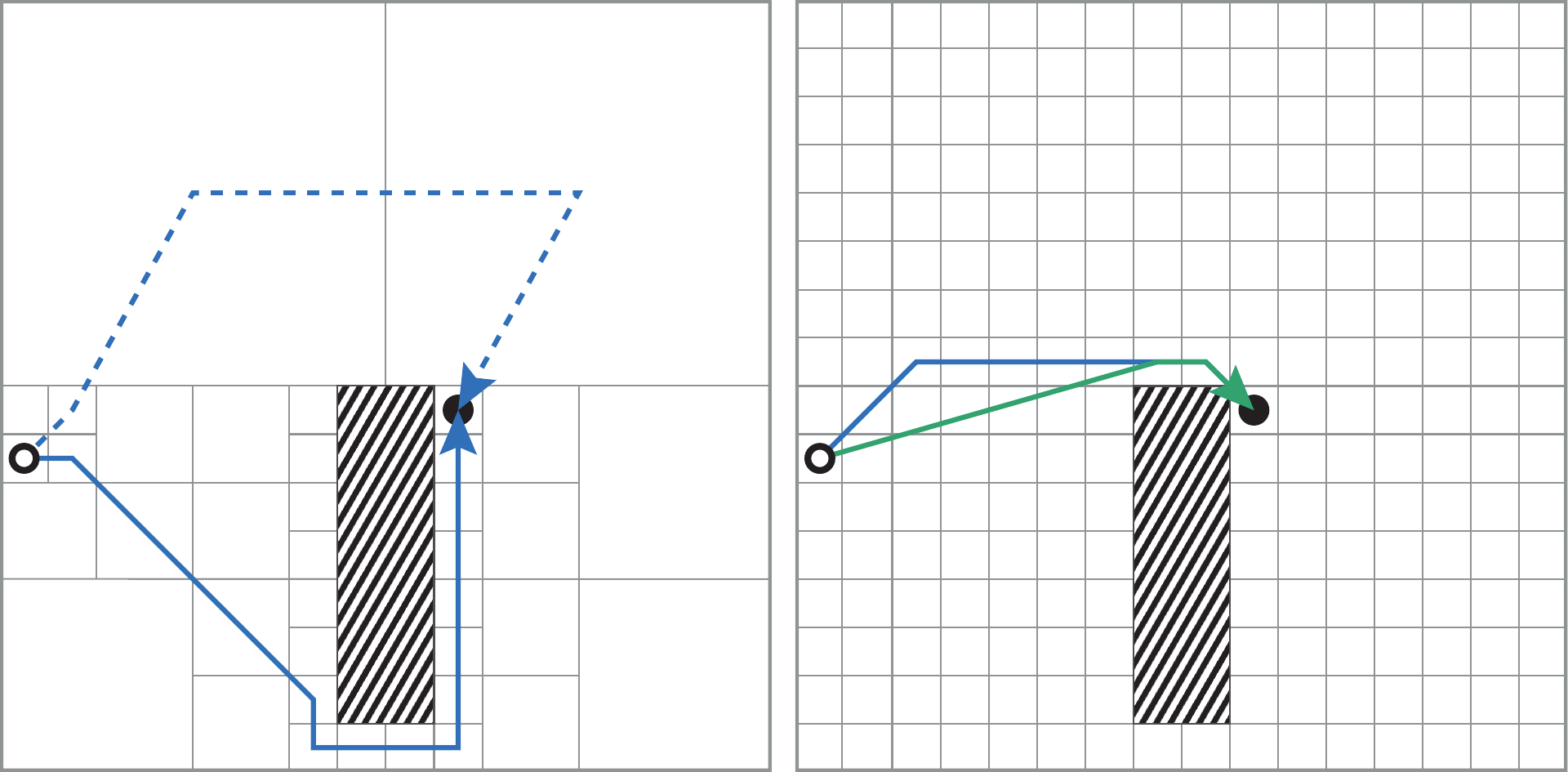}
    \caption{Comparison of A* on an octree's leaves (left, {\color[HTML]{0073B2}blue}), A* on a fixed-resolution grid (right, {\color[HTML]{0073B2}blue}), and Theta* (right, {\color[HTML]{009E73}green}). On the octree, A* produces highly suboptimal paths. While its search space includes a path (dashed {\color[HTML]{0073B2}blue}) on the correct side of the obstacle (striped box), this path is ignored due to the detour introduced by passing through the leaves' centers. A* on the grid finds shorter, smoother paths, but still performs worse than Theta*. Our method matches Theta*'s path quality while operating on octrees.}
    \label{fig:global_planning/media/octree_a_star_wrong_homotopy}
\end{figure}

Octree-based maps~\cite{hornungOctoMapEfficientProbabilistic2013,vespaEfficientOctreeBasedVolumetric2018,reijgwart2023wavemap} compactly encode traversability information using multi-resolution. Evaluating A* directly on the adjacency graph of an octree's leaves preserves the completeness of running it on a grid at the highest resolution while offering significant memory and runtime improvements~\cite{Kambhampati1986OctreeAStar}. However, considering only the octree leaves' centers yields suboptimal paths in length and smoothness~\cite{Chen1997framedQuadtree}. As illustrated in \Cref{fig:global_planning/media/octree_a_star_wrong_homotopy}, post-processing steps such as path-shortening cannot resolve this issue since the paths might not even be close to the true shortest path.

Any-angle planning algorithms, such as Theta*~\cite{daniel2010ThetaStar}, improve path quality by allowing paths to deviate from grid edges, connecting vertices in line of sight with straight lines. This approach can yield paths up to $\approx13\%$ shorter than those produced by A*~\cite{nash2010LazyThetaStar}. In this paper, we extend Theta*'s cost field formulation to efficiently represent large parts of the search space at coarser resolutions. Additionally, we propose a coarse-to-fine search algorithm that starts at the coarsest resolution and refines solutions only in regions requiring higher detail, significantly reducing memory usage and computational cost without sacrificing accuracy.

In summary, the main contribution of this paper is a search-based planner that combines the accuracy of any-angle planning with the efficiency of multi-resolution representations and hierarchical algorithms. Extensive evaluations on synthetic and real-world maps demonstrate that the proposed method retains Theta*'s accuracy while running up to two orders of magnitude faster. Compared to well-established search- and sampling-based planners, it consistently finds near-optimal paths and, in cluttered environments, runs even faster than sampling-based planners. The complete framework is open-sourced\footnote{\url{https://github.com/ethz-asl/wavestar}} to allow the planning and robotics communities to build on these results.

%% file: sections/related_work.tex
\section{Related work}
\label{sec:global_planning/related_work}
Path planning methods can generally be categorized into optimization-, sampling-, and search-based approaches. Sampling-based methods are most commonly used for global planning, especially in large environments. While very fast, randomized methods such as RRT~\cite{lavalle2006planningAlgorithms} and RRTConnect~\cite{kuffner2000RRTConnect} provide no guarantees on the quality of their solutions. Variants such as RRT*~\cite{karaman2011RRTStar} are guaranteed to converge the optimal solution as the number of samples approaches infinity. However, they do not provide bounds on their intermediate solutions and stopping them after a finite time leads to different paths even when the start and goal positions are the same~\cite{Du2020MRAStar}. A challenge in practice is that the inconsistency of randomized planners worsens in cluttered environments, and finding solutions through narrow passages can take a very long time.

Search-based planners such as A*~\cite{hart1968Astar} operate on a discretized search space and are deterministic, complete, and terminate in finite time, explicitly reporting when no solution exists. However, using a fixed-resolution 3D occupancy map's adjacency graph as the space discretization results in runtimes that grow linearly with volume and cubically with resolution, making it impractical for large or high-resolution maps.

Several research efforts have explored hierarchical approaches to improve the scalability of search-based planning. Kambhampati and Davis~\cite{Kambhampati1986OctreeAStar} applied A* to the octree's leaves, compactly representing traversable space and achieving significant efficiency improvements, albeit at the cost of longer, jagged paths. Funk et al.~\cite{funk2023orientationAStar} extended this approach to orientation-aware planning in large environments with narrow openings. CFA*~\cite{lee2009CFAStar} proposed a coarse-to-fine strategy, performing a coarse search over large blocks followed by refinement at the grid cell level. HPA*~\cite{botea2004HPAStar} generalized this concept to multi-level hierarchies of pre-processed clusters. Beyond these methods, iterative~\cite{hauer2015multi} and information-theoretic~\cite{larsson2021information} approaches have also been proposed. Recently, Du et al.~\cite{Du2020MRAStar} demonstrated how multiple simultaneous weighted-A* searches at different resolution levels can share information to combine their strengths. However, a significant drawback of all these methods is that their path lengths are, at best, equivalent to those of A* on the highest-resolution grid.

Any-angle planners improve upon A* by allowing deviations from the grid's edges, finding up to $\approx13\%$ shorter paths~\cite{nash2010LazyThetaStar} by better approximating true shortest paths in continuous space, which are \textit{taut} -- straight except at inflection points wrapping around obstacles. Theta*~\cite{daniel2010ThetaStar}, a widely adopted any-angle planner, achieves high accuracy in diverse environments~\cite{uras2021anyAngleComparison} by connecting each vertex to its best visible predecessor. While these deviations improve accuracy, Theta* propagates information only along grid edges, enabling simple and efficient implementation. However, in 3D, it incurs significant runtime overhead due to the numerous visibility checks required to ensure vertex-predecessor edges are collision-free. LazyTheta*~\cite{nash2010LazyThetaStar} addresses this limitation with lazy visibility checking, reducing the overhead by an order of magnitude with minimal impact on path quality.

Multi-resolution methods for any-angle planning have also been explored. Chen et al.~\cite{Chen1997framedQuadtree} introduced framed quadtrees, which pad leaf nodes with high-resolution vertices to permit a broader range of angles through each leaf. While effective in 2D, this approach scales poorly for 3D Euclidean shortest paths, multiplying A*'s computational complexity by an additional term that grows quartically with padding resolution. Closest to our work, Faria et al.~\cite{faria2019efficient} applied LazyTheta* to octree leaves. However, their method produces arbitrarily suboptimal paths as it considers only leaf centers. In contrast, we explicitly consider the high-resolution vertices within each leaf and dynamically refine the octree to bound the approximation error. Additionally, a custom initialization procedure ensures all potentially optimal inflection points are evaluated. Extensive comparisons and ablations demonstrate that these improvements yield significantly shorter, smoother paths.

%% file: sections/problem_statement.tex
\section{Problem statement}
\label{sec:problem_statement}
This paper presents a method for finding collision-free Euclidean shortest paths between a start and goal point in 2D or 3D workspaces, given an occupancy map representing obstacles. To simplify collision checking, we approximate the robot as a bounding sphere and inflate all obstacles by its radius, treating the robot as a point. Like Theta*, the presented algorithm does not account for motion constraints.

%% file: sections/method.tex
\section{Method}
In this section, we describe the components of our planner. First, we show how multi-resolution enables compact encoding of intermediate solutions in any-angle planning. Building on this, we present our efficient any-angle path planner, which consists of: i) an approach that ensures all plausible waypoints are efficiently considered, and ii) a coarse-to-fine method to explore the search space spanned by the previously generated waypoints. Formal statements on our method's completeness, optimality and time-complexity are provided in \Cref{appendix:formal_statements}.

\subsection{Any-angle planning}
Any-angle planners produce shorter and smoother paths than A* by allowing paths to deviate from grid edges (\Cref{fig:global_planning/media/octree_a_star_wrong_homotopy} right), better approximating true shortest paths in continuous space. A necessary condition for Euclidean shortest paths is that they are \textit{taut}, i.e. consisting of straight line segments connected at inflection points where the path wraps tightly around obstacles. 

We base our planner on Theta*~\cite{daniel2010ThetaStar}, which closely follows A*’s algorithm. Both planners check whether a path through an expanded node can improve the cost-to-come ($g$ cost) of its neighbors. However, Theta* introduces a key improvement: for each neighbor, it performs a visibility check to determine if it can be directly connected to the node's $\Predecessor$. As illustrated in \Cref{fig:global_planning/media/g_cost_vs_predecessor_field}, this eliminates intermediate waypoints, further reducing $g$ costs and avoiding detours.

\subsection{Multi-resolution cost field representation}
\label{subsec:global_planning/multi_resolution_cost_field}
Search-based planners such as A* and Theta* compute the minimum $g$ cost and best $\Predecessor$ for each (grid) vertex expanded during the search. Since these two properties are often stored together, we refer to their combination as the \emph{cost field}.

\begin{figure}[bt]
    \centering
    \includegraphics[width=\linewidth]{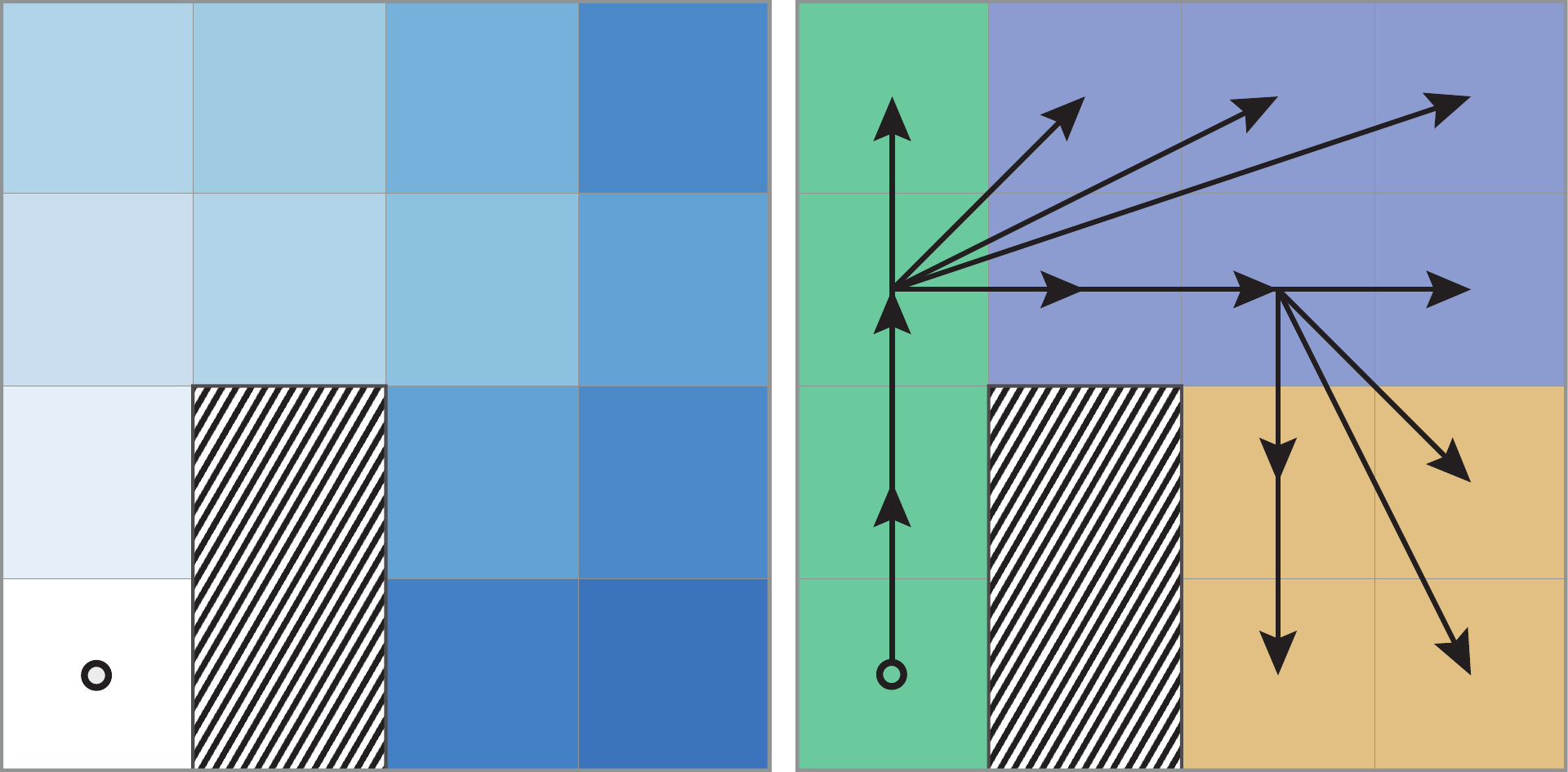}
    \caption{Illustration of the cost-to-come ($g$ cost) and $\Predecessor$ fields of Theta* in a 2D environment with a single obstacle (striped box). The $g$ cost field (left) changes from cell to cell, while the $\Predecessor$ field (right) is largely constant. All cells to the left of the obstacle ({\color[HTML]{63CBA9}green}) are directly visible from the start vertex (black circle) and thus use it as their predecessor. Cells near the top right ({\color[HTML]{82A7D1}purple}) connect through the cell at the obstacle's top-left corner, while the rest ({\color[HTML]{EAC595}gold}) connect through the cell at its top-right corner.}
    \label{fig:global_planning/media/g_cost_vs_predecessor_field}
\end{figure}

By construction, neighboring grid vertices rarely share the same $g$ cost. In Theta*, however, large regions are often dominated by the same $\Predecessor$ (\cref{fig:global_planning/media/g_cost_vs_predecessor_field}). The $g$ cost of any vertex $s$ is defined as the $g$ cost of its predecessor plus the distance between them. Consequently, the cost field can be compressed losslessly by storing $\Predecessor(\mathcal{V})$ and $g(\Predecessor(\mathcal{V}))$ for each region $\mathcal{V}$. The $g$ cost of any vertex $s$ within a region $\mathcal{V}$ can then be retrieved using
\begin{equation}
    \label{eq:global_planning/cost_field_cell_g_cost_from_volume}
    g(s) = g(s^p) + c(s^p, s)\; \big|\; s\in\mathcal{V}, s^p=\Predecessor(\mathcal{V})
\end{equation}
where $c(s^p, s)$ is the Euclidean distance from $s^p$ to $s$.

To exploit this, we propose to partition the cost field into multi-resolution cubes corresponding to an octree's leaves. This regular structure enables efficient storage, fast random access, simplified neighborhood operations, and natural alignment between the search space and octree-based traversability maps.

\subsection{Cost field initialization}
\label{subsec:global_planning/initializing_inflection_points}
Just as typical search-based planners where a vertex's predecessor is itself a vertex, we define the predecessors of subvolumes as subvolumes. While a single subvolume could encompass the start, goal, and several inflection points, we initialize the cost field such that each subvolume contains at most one such waypoint to reduce bookkeeping and the runtime complexity of neighborhood operations.

\begin{figure}[bt]
    \centering
    \includegraphics[width=\linewidth]{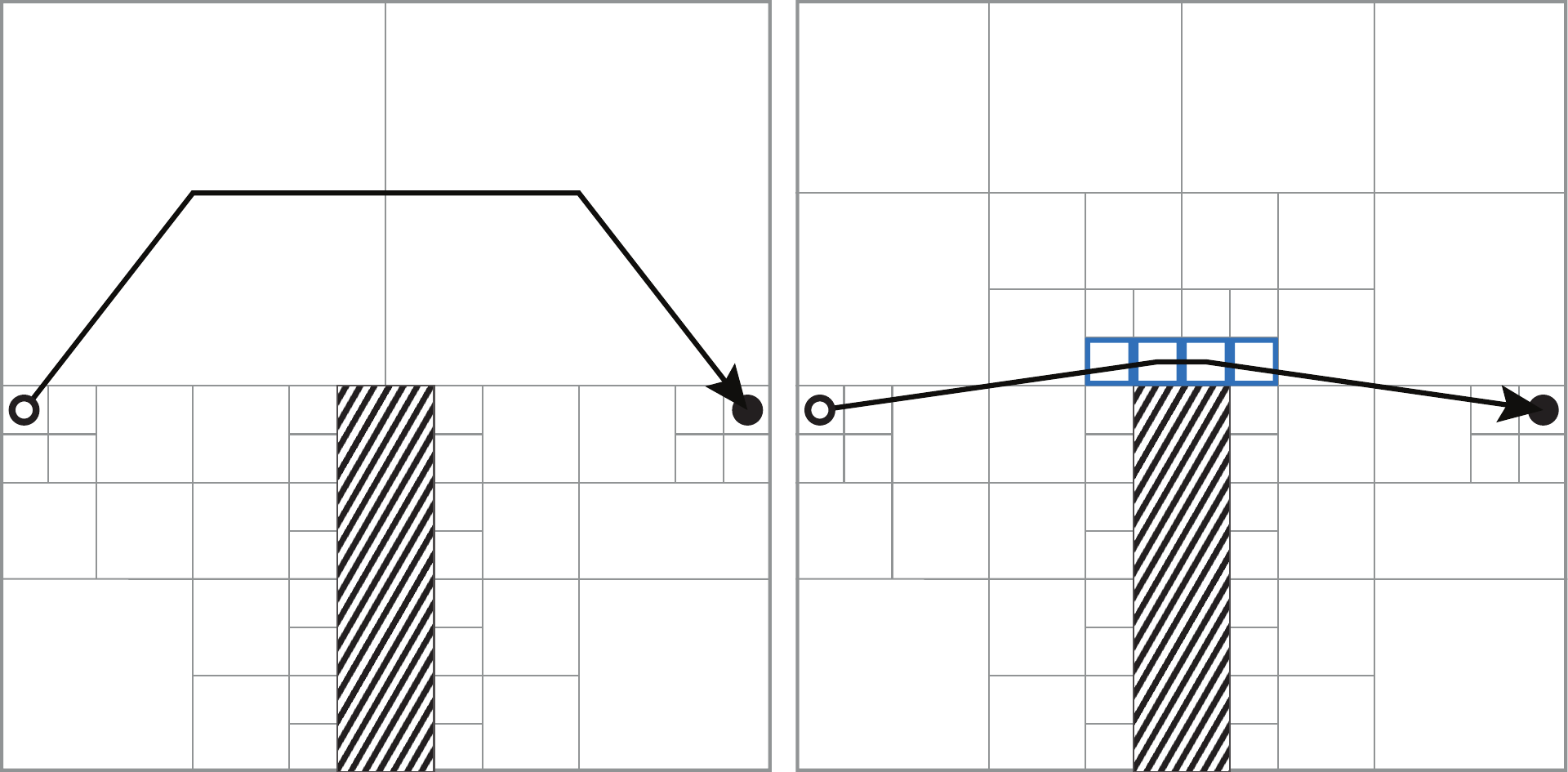}
    \caption{Illustration of the importance of initializing inflection points. Without initialization (left), the retrieved shortest path may take large detours around obstacles. Initializing the cost field at a higher resolution near obstacles (right) resolves this issue, resulting in shorter, smoother paths. As the number of added subvolumes ({\color[HTML]{0073B2}blue}) is small, the performance overhead remains minimal.}
    \label{fig:global_planning/media/initialization}
\end{figure}

Initializing subvolumes at the minimum resolution required to avoid occupied leaves in the map's octree suffices to guarantee resolution completeness~\cite{Kambhampati1986OctreeAStar}. However, as shown in \Cref{fig:global_planning/media/initialization} (left), this approach often results in paths taking significant detours around obstacles. From the definition of taut paths, optimal inflection points can only appear next to obstacles. Thus, considering all vertices of a high-resolution grid that are traversable and adjacent to an obstacle ensures that no inflection points that would be considered by Theta* on the same grid are missed. This initialization can be performed globally or incrementally as the search progresses through the traversability map. In our tests, we adopt the incremental approach, allowing the initialization cost to scale with the explored volume rather than the map's total size.

\subsection{Multi-resolution search}
\label{subsec:global_planning/multi_resolution_search}
Search-based planning in 3D spaces is computationally challenging as the number of vertices grows rapidly with increasing resolution. To address this, our approach explores the search space at the coarsest possible resolution and dynamically refines it only where needed to maintain accuracy. After initializing the cost field as described in \cref{subsec:global_planning/initializing_inflection_points}, our any-angle planner computes each subvolume's $\Predecessor(\mathcal{V})$ and $g(\Predecessor(\mathcal{V}))$ by running a modified version of A* over the octree's leaves. Similar to A*, the algorithm uses a min-priority queue ($\OpenQueue$) to expand elements sorted by their estimated goal-reaching cost ($f$ score). However, unlike A*, the elements in our algorithm are subvolumes that can contain many vertices, which are processed together.

\begin{algorithm}[bt]
    \SetAlgoLined
    \SetNoFillComment
    \DontPrintSemicolon

    \SetKwProg{Fn}{Function}{ is}{end}
    \SetKwFunction{UpdateSubvolume}{UpdateSubvolume}
    \SetKwFunction{ComputeFScore}{ComputeFScore}
    \SetKwFunction{CheckNode}{CheckNode}

    $\OpenQueue \gets \emptyset$\;
    $\ClosedSet \gets \emptyset$\;
    $g(s^\text{start}) \gets 0$\;
    $\Predecessor(\mathcal{V}^\text{start}) \gets s^\text{start}$\;
    $\OpenQueue.insert(\mathcal{V}^\text{start}, \ComputeFScore(\mathcal{V}^\text{start}))$\;
    \While{$\OpenQueue \neq \emptyset$}{%
        $\mathcal{V} \gets \OpenQueue.pop()$\;\label{algoline:global_planning/multi_res_heuristic_guided_graph_search/expand_v}
        \If{$s^\text{goal} \in \mathcal{V}$}{\label{algoline:global_planning/multi_res_heuristic_guided_graph_search/goal_reached}%
            \Return{\PathFound}
        }
        $\ClosedSet \gets \ClosedSet \cup \{\mathcal{V}\}$\;
        $g(\mathcal{V}_\text{center}) \gets g(\Predecessor(\mathcal{V})) + c(\Predecessor(\mathcal{V}), \mathcal{V}_\text{center})$\; \label{algoline:global_planning/multi_res_heuristic_guided_graph_search/set_g_cost_Vcenter}
        $\UpdateSubvolume(\mathcal{V}, \mathcal{V}_\text{root})$\; \label{algoline:global_planning/multi_res_heuristic_guided_graph_search/start_recursion}
    }
    \Return{\NoPathFound}
    
    \caption{Heuristic-guided search over subvolumes}
    \label{algo:global_planning/multi_res_heuristic_guided_graph_search}
\end{algorithm}

\Cref{algo:global_planning/multi_res_heuristic_guided_graph_search} shows our planner's main loop. For each expanded subvolume $\mathcal{V}$, the algorithm first checks whether it contains the goal vertex $s^\text{goal}$ (\cref{algoline:global_planning/multi_res_heuristic_guided_graph_search/goal_reached}). If so, the search terminates. Otherwise, $\mathcal{V}$ is added to the $\ClosedSet$ set, and the $g$ cost for its center is computed (\cref{algoline:global_planning/multi_res_heuristic_guided_graph_search/set_g_cost_Vcenter}) and stored for potential use as a $\Predecessor$. Finally, the $\UpdateSubvolume$ function processes all adjacent subvolumes, which may include 26 or more multi-resolution neighbors.

\subsection{Dynamic refinement}
\label{subsec:global_planning/dynamic_refinement}
\begin{figure}[bt]
    \centering
    \includegraphics[width=\linewidth]{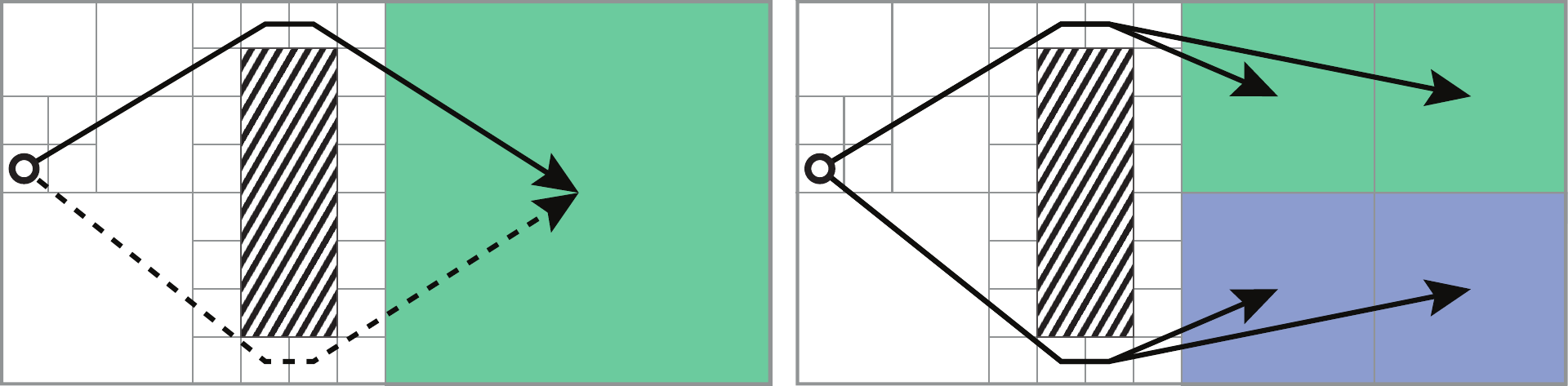}
    \caption{Illustration of our dynamic refinement procedure. When a new path (dashed arrow) is discovered to a subvolume $\mathcal{V}'$ (left) that has already been reached (solid arrow), we evaluate its effect on the cost to reach each vertex in $\mathcal{V}'$. The algorithm handles three cases: i) the new path reduces the cost for all vertices, replacing the previous path; ii) the new path does not improve any costs and is ignored; iii) some costs improve while others worsen. In this last case (right), $\mathcal{V}'$ is recursively subdivided until each child subvolume is fully resolved under case i or ii. Subvolumes are colored by their $\Predecessor$.}
    \label{fig:global_planning/media/refinement}
\end{figure}

Throughout the majority of the environment, the cost field resolution chosen by the initialization procedure suffices. However, subvolumes are occasionally reachable from multiple predecessors, each better suited for different vertices within the subvolume. For instance, in \Cref{fig:global_planning/media/refinement}, vertices toward the top right are optimally reached by passing above the obstacle, whereas passing below the obstacle provides shorter paths to the vertices on the bottom right. In such cases, instead of assigning a single suboptimal $\Predecessor$ to the entire subvolume, we dynamically refine the cost field.

\begin{algorithm}[bt]
    \SetAlgoLined
    \SetNoFillComment
    \DontPrintSemicolon

    \SetKwProg{Fn}{Function}{ is}{end}
    \SetKwFunction{UpdateSubvolume}{UpdateSubvolume}
    \SetKwFunction{UpdateCost}{UpdateCost}
    \SetKwFunction{ComputeFScore}{ComputeFScore}
    \SetKwFunction{CheckNode}{CheckNode}
    \SetKwFunction{IsLeaf}{IsLeaf}
    \SetKwFunction{AreAdjacent}{AreAdjacent}
    \SetKwFunction{LineOfSight}{LineOfSight}

    \Fn{$\UpdateSubvolume(\mathcal{V}, \mathcal{V}')$}{%
        \If{$\IsLeaf(\mathcal{V}')$}{ \label{algoline:global_planning/multi_res_recursive_update_node/is_leaf}%
            Status $\gets \UpdateCost(\mathcal{V}, \mathcal{V}')$\;
            \uIf{Status = \StrictlyBetter}{%
                \If{$\mathcal{V}' \in \OpenQueue$}{%
                    $\OpenQueue.remove(\mathcal{V}')$\;
                }
                $\OpenQueue.insert(\mathcal{V}', \ComputeFScore(\mathcal{V}'))$\; \label{algoline:global_planning/multi_res_recursive_update_node/update_priority_in_open}
                \Return\;
            }
            \uElseIf{Status = \NotBetter}{%
                \Return\;
            }
            \Else(\tcp*[h]{Status = \Ambiguous}){%
                $\OpenQueue.remove(\mathcal{V}')$\; \label{algoline:global_planning/multi_res_recursive_update_node/remove_parent_from_open}
            }
        }
        
        \ForEach{$\mathcal{V}^{'\text{child}} \in \mathcal{V}'$}{ \label{algoline:global_planning/multi_res_recursive_update_node/for_loop}%
            \If{$\mathcal{V}^{'\text{child}} \notin \ClosedSet$}{%
                $s^{p'} \gets \Predecessor(\mathcal{V}')$\;
                \If{$\mathcal{V}^{'\text{child}} \notin \OpenQueue$}{%
                    $\Predecessor(\mathcal{V}^{'\text{child}}) \gets s^{p'}$\;
                    $\OpenQueue.insert(\mathcal{V}^{'\text{child}}, \ComputeFScore(\mathcal{V}^{'\text{child}}))$
                }
                \If{$\AreAdjacent(\mathcal{V}, \mathcal{V}^{'\text{child}})$
                    $\;\textbf{and}\; \mathcal{V}^{'\text{child}} \neq \mathcal{V}$}{ \label{algoline:global_planning/multi_res_recursive_update_node/recurse_if_adjacent}%
                    $\UpdateSubvolume(\mathcal{V}, \mathcal{V}^{'\text{child}})$\;
                }
            }
        }
    }
    \BlankLine
    \Fn{$\AreAdjacent(\mathcal{V}^a,\mathcal{V}^b)$}{
        $\mathbf{sep} \gets $
        $\text{cmax}(\mathcal{V}^a_{\min}, \mathcal{V}^b_{\min})$
        $-$
        $\text{cmin}(\mathcal{V}^a_{\max}, \mathcal{V}^b_{\max})$\;
        \Return{$\mathbf{sep}.x \leq 1 \;\textbf{and}\; \mathbf{sep}.y \leq 1 \;\textbf{and}\; \mathbf{sep}.z \leq 1$}
    }
    \caption{Recursive octree node updates}
    \label{algo:global_planning/multi_res_recursive_update_node}
\end{algorithm}

Leveraging the hierarchical structure of octrees, this refinement can efficiently be integrated into our multi-resolution planner through a recursive $\UpdateSubvolume$ method (\cref{algo:global_planning/multi_res_recursive_update_node}). Starting at the octree's root $\mathcal{V}' = \mathcal{V}_\text{root}$, the method handles two cases. If $\mathcal{V}'$ is a leaf, the function calls $\UpdateCost$ to evaluate whether a path through the expanded subvolume $\mathcal{V}$ or its $\Predecessor(\mathcal{V})$ improves the $g$ cost of $\mathcal{V}'$. Since $\mathcal{V}'$ can contain multiple vertices, the comparison has three possible outcomes: i) using $\mathcal{V}$ or $\Predecessor(\mathcal{V})$ reduces the cost for all vertices in $\mathcal{V}'$, updating its $\Predecessor$ and reprioritizing it in the $\OpenQueue$ queue; ii) no costs improve and both predecessors are ignored; or iii) some costs improve while others worsen, flagging $\mathcal{V}'$ for refinement.

In the second case, for non-leaf subvolumes or those flagged for refinement, the function iterates over $\mathcal{V}'$'s children, skipping $\ClosedSet$ nodes. Newly created children inherit their parent's $\Predecessor$ and are added to the $\OpenQueue$ queue, ensuring the full region $\mathcal{V}'$ covered is eventually processed. The function then recursively visits each child adjacent to $\mathcal{V}$, until all descendant subvolumes fall under cases 1 or 2.
$\AreAdjacent$ tests whether subvolumes $\mathcal{V}^a$ and $\mathcal{V}^b$ touch or overlap, which holds if the minimum offset between their \acp{AABB} is at most 1 along every axis on the highest-resolution grid. This offset is computed from the coefficient-wise max ($\text{cmax}$) and min ($\text{cmin}$) of their \ac{AABB} corners $\mathcal{V}^{\cdot}_{\min}$ and $\mathcal{V}^{\cdot}_{\max}$.

The $\UpdateCost$ function (\cref{algo:global_planning/multi_res_theta_star}) implements the comparison between $\mathcal{V}'$'s current $\Predecessor$, $s^{p'}$, and two candidate predecessors: an inflection point at the center of $\mathcal{V}$, $s^c$, and $\Predecessor(\mathcal{V})$, $s^p$. If $s^{p'}$ results in a lower $g$ cost for all vertices $s \in \mathcal{V}'$ compared to $s^p$ and $s^c$, the function returns $\NotBetter$, indicating no changes are required. Conversely, if $s^p$ or $s^c$ provides strictly better costs for all $s \in \mathcal{V}'$, the $\Predecessor$ is updated and $\StrictlyBetter$ is returned. If neither condition is fully satisfied, the function returns $\Ambiguous$, signaling the need for further refinement.

In practice, tolerating small path length suboptimalities may be acceptable, particularly if it leads to efficiency improvements. To quantify this, we define the worst-case suboptimality of $s^{p'}$ relative to an alternative predecessor $s^i$ over $\mathcal{V}'$ as
\begin{align}
    \label{eq:global_planning/cost_field_approximation_error}
    \hat{E}\big(s^{p'}, s^i, \mathcal{V}'\big)
    &= \max_{s \in \mathcal{V}'} \frac{g(s^{p'}) + c(s^{p'}, s) - g(s^i) - c(s^i, s)}{c(s^{p'}, s)}
\end{align}
where $c(s^a, s^b)$ is the straight-line distance from $s^a$ to $s^b$. Since the error is normalized by edge length, the total accumulated error along the path grows at most proportionally with the path's length.

To bound this error, we introduce $\epsilon$, which represents the worst-case relative path length suboptimality. The function $\IsBetterOrSimilar$ (\cref{algoline:global_planning/multi_res_theta_star/is_better_or_similar_def}) applies this threshold to decide whether a new predecessor should be accepted. For $\epsilon = 0$, the function returns $\texttt{True}$ only if $s^a$ is a strictly better predecessor than $s^b$ for every vertex in $\mathcal{V}'$. The planner then refines each subvolume until its children are strictly dominated by a single predecessor. In general, $\UpdateSubvolume$ recurses until the following condition holds for $s^i\in\{s^p,s^c\}$:
\begin{align}
    \label{eq:global_planning/cost_field_approximation_error_invariant}
    \hat{E}\big(\Predecessor\big(\mathcal{V}'\big), s^i, \mathcal{V}'\big) \leq \epsilon
\end{align}

Finally, $\ComputeFScore$ demonstrates how our multi-resolution planner computes consistent $f$ scores for sorting the $\OpenQueue$ queue, by identifying the minimum $f$ score across all vertices in $\mathcal{V}$. Although $\IsBetterOrSimilar$ and $\ComputeFScore$ consider subvolumes with many vertices, the computational burden is significantly reduced in practice because $c(s^p, s)$ and $h(s)$ are straight-line distances, requiring only a few critical vertices to be checked.

\begin{algorithm}[bt]
    \SetAlgoLined
    \SetNoFillComment
    \DontPrintSemicolon

    \SetKwProg{Fn}{Function}{ is}{end}
    \SetKwFunction{UpdateSubvolume}{UpdateSubvolume}
    \SetKwFunction{UpdateCost}{UpdateCost}
    \SetKwFunction{ComputeFScore}{ComputeFScore}
    \SetKwFunction{CheckNode}{CheckNode}
    \SetKwFunction{IsLeaf}{IsLeaf}
    \SetKwFunction{AreAdjacent}{AreAdjacent}
    \SetKwFunction{LineOfSight}{LineOfSight}
    \SetKwFunction{IsBetterOrSimilar}{IsBetterOrSimilar}

    \Fn{$\UpdateCost(\mathcal{V}, \mathcal{V}')$}{%
        $s^c \gets \mathcal{V}_\text{center}$\;
        $s^p \gets \Predecessor(\mathcal{V})$\;
        $s^{p'} \gets \Predecessor(\mathcal{V}')$\;
        \eIf{$\LineOfSight(s^p, \mathcal{V}')$}{\label{algoline:global_planning/multi_res_theta_star/has_line_of_sight}%
            \tcp{Ray traced connection}
            \uIf{$\IsBetterOrSimilar(s^{p'}, s^p, \mathcal{V}')$}{%
                \Return{\NotBetter}\;
            }
            \ElseIf{$\IsBetterOrSimilar(s^p, s^{p'}, \mathcal{V}')$}{%
                $\Predecessor(\mathcal{V}') \gets s^p$\;
                \Return{\StrictlyBetter}\;
            }
        }{%
            \tcp{Direct neighbor connection}
            \uIf{$\IsBetterOrSimilar(s^{p'}, s^c, \mathcal{V}')$}{%
                \Return{\NotBetter}\;
            }
            \ElseIf{$\IsBetterOrSimilar(s^c, s^{p'}, \mathcal{V}')$}{%
                $\Predecessor(\mathcal{V}') \gets s^c$\;
                \Return{\StrictlyBetter}\;
            }
        }
        \Return{\Ambiguous}\;
    }
    \BlankLine
    \Fn{$\IsBetterOrSimilar(s^a, s^b, \mathcal{V}')$}{ \label{algoline:global_planning/multi_res_theta_star/is_better_or_similar_def}%
        \eIf{$\forall s \in \mathcal{V}':$
            $g(s^a) + c(s^a, s)$
            $<$
            $ g(s^b) + c(s^b, s) + \epsilon\ c(s^a, s) $}{%
            \Return{True}\;
        }{%
            \Return{False}\;
        }  
    }
    \BlankLine
    \Fn{$\ComputeFScore(\mathcal{V})$}{
        $s^p \gets \Predecessor(\mathcal{V})$\;
        \Return{$\min_{s \in \mathcal{V}} \left[ g(s^p) + c(s^p, s) + h(s) \right]$}\;
    }
    \caption{Computing cost updates and heuristics}
    \label{algo:global_planning/multi_res_theta_star}
\end{algorithm}

%% file: sections/experiments.tex
\section{Experiments}
In the experiments, we start with an evaluation of our multi-resolution planner's initialization and dynamic refinement procedures. Then, we compare our method to other path-planning approaches based on success rate, path length, and runtime.

The result plots share an overall structure, showing the measured quantity on the Y-axis, while the property being varied is along the X-axis. The result distributions are presented as box plots with individual results shown as dots, colored according to the true path length. Where results are relative to a baseline, a blue line indicates baseline performance.

\subsection{Impact of initialization and refinement}
\label{subsec:global_planning/ablations}
The evaluation of our initialization and refinement procedures impact on performance are conducted on five synthetic maps, each measuring $\SI{100}{\meter} \times \SI{100}{\meter} \times \SI{100}{\meter}$ mapped at $\SI{10}{\centi\meter}$ resolution. They are generated by adding $0$, $1000$, $2000$, $3000$, and $4000$ randomly shaped obstacles to an initially empty volume, representing varying levels of clutter. For each map, $100$ random collision-free start-goal pairs ($500$ total) were sampled.

\subsubsection{Inflection point initialization}
To evaluate the significance of the initialization procedure (\cref{subsec:global_planning/initializing_inflection_points}), we compare the path lengths and execution times of our planner with and without initialization to $\ThetaStar$ running at the highest resolution. As our method allows defining the initialization resolution, we present results for initialization resolutions ranging from $\SI{1.6}{\meter}$ to $\SI{10}{\centi\meter}$. For this analysis, the dynamic refinement strategy (\cref{subsec:global_planning/dynamic_refinement}) is disabled to isolate the effect of initialization.

\begin{figure}[bt]
    \centering
    \includegraphics[width=\linewidth]{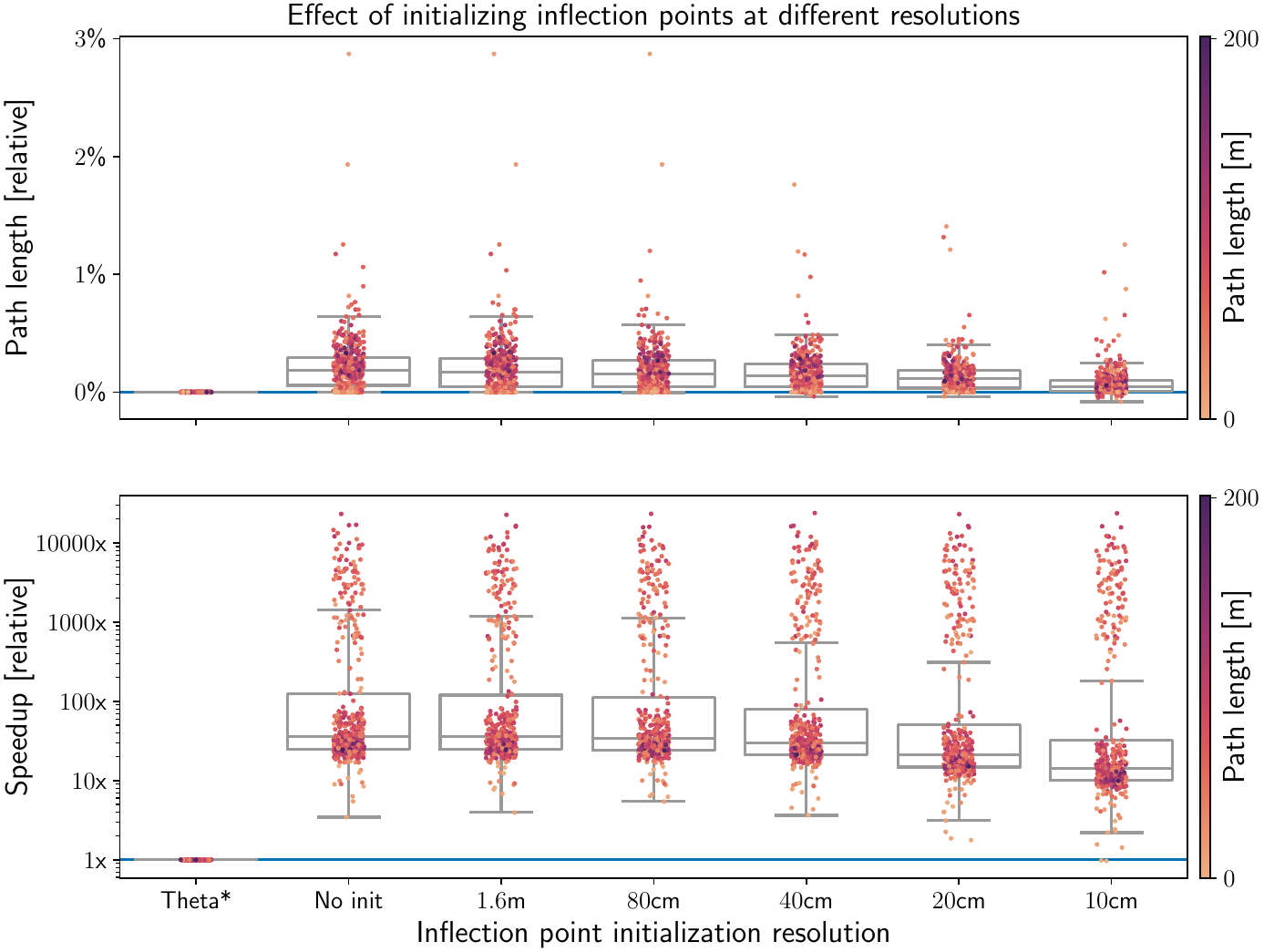}
    \caption{Initialization procedure's impact on path length and runtime relative to Theta* ({\color[HTML]{0073B2}blue} line). Reading the plot from left to right, we see how increasing the inflection point initialization resolution brings our path lengths (top) closer to Theta* while remaining significantly faster (bottom).}
    \label{fig:global_planning/ablations_seeding}
\end{figure}

The results in \Cref{fig:global_planning/ablations_seeding} (top) show that increasing the inflection point initialization resolution moves the path lengths of our planner closer to those of $\ThetaStar$. On average, path lengths converge to values close to $\ThetaStar$, especially for longer paths (deep purple). Outliers primarily correspond to short paths (light orange), where small differences are amplified when normalized by the short path length and runtime values of $\ThetaStar$.

At coarser resolutions ($\geq$\SI{80}{\centi\meter}), initializing inflection points yields no notable improvement over $\NoExplicitInit$ because the cost field's octree conforms to the obstacles in the occupancy map. This results in most occupied cells being surrounded by medium to high-resolution subvolumes, rendering low-resolution initializations redundant. Additionally, the occupancy map and cost field are stored using an optimized octree data structure~\cite{musethOpenVDBOpensourceData2013}, limiting the coarsest resolution to \SI{6.4}{\meter}, implicitly preventing extremely bad solutions. At higher resolutions, our method occasionally discovers slightly shorter paths than $\ThetaStar$, as $\ThetaStar$ is not guaranteed to be optimal.

Looking at runtime results (\Cref{fig:global_planning/ablations_seeding} bottom) we see that increasing the initialization resolution generally increases runtime, as higher resolutions require the search to expand more subvolumes and consider more inflection points as predecessors. Note that the clusters of points at the top of the bottom plot correspond to queries in the $0$-obstacle environment. Since these involve no obstacles requiring initialization, their speedup is independent of the initialization resolution.

\subsubsection{Refinement strategy}
To evaluate the effect of the dynamic refinement procedure (\cref{subsec:global_planning/dynamic_refinement}), we run our planner with different approximation error thresholds $\epsilon$ (\cref{eq:global_planning/cost_field_approximation_error_invariant}) and compare the results to $\ThetaStar$. The initialization procedure (\cref{subsec:global_planning/initializing_inflection_points}) is disabled to isolate the effect of refinement. We test $\epsilon$ values ranging from $10^{-1}$ to $10^{-3}$, along with two special cases: lossless refinement ($\epsilon=0$) and no refinement, which we refer to as $\MatchMap$. Because the planner can only traverse fully unoccupied cells, the resolution of the cost field must always match or exceed the occupancy map's leaf resolution. When refinement and initialization are disabled, the cost field's subvolumes exactly match the occupancy map's leaves, resulting in the $\MatchMap$ configuration.

\begin{figure}[bt]
    \centering
    \includegraphics[width=\linewidth]{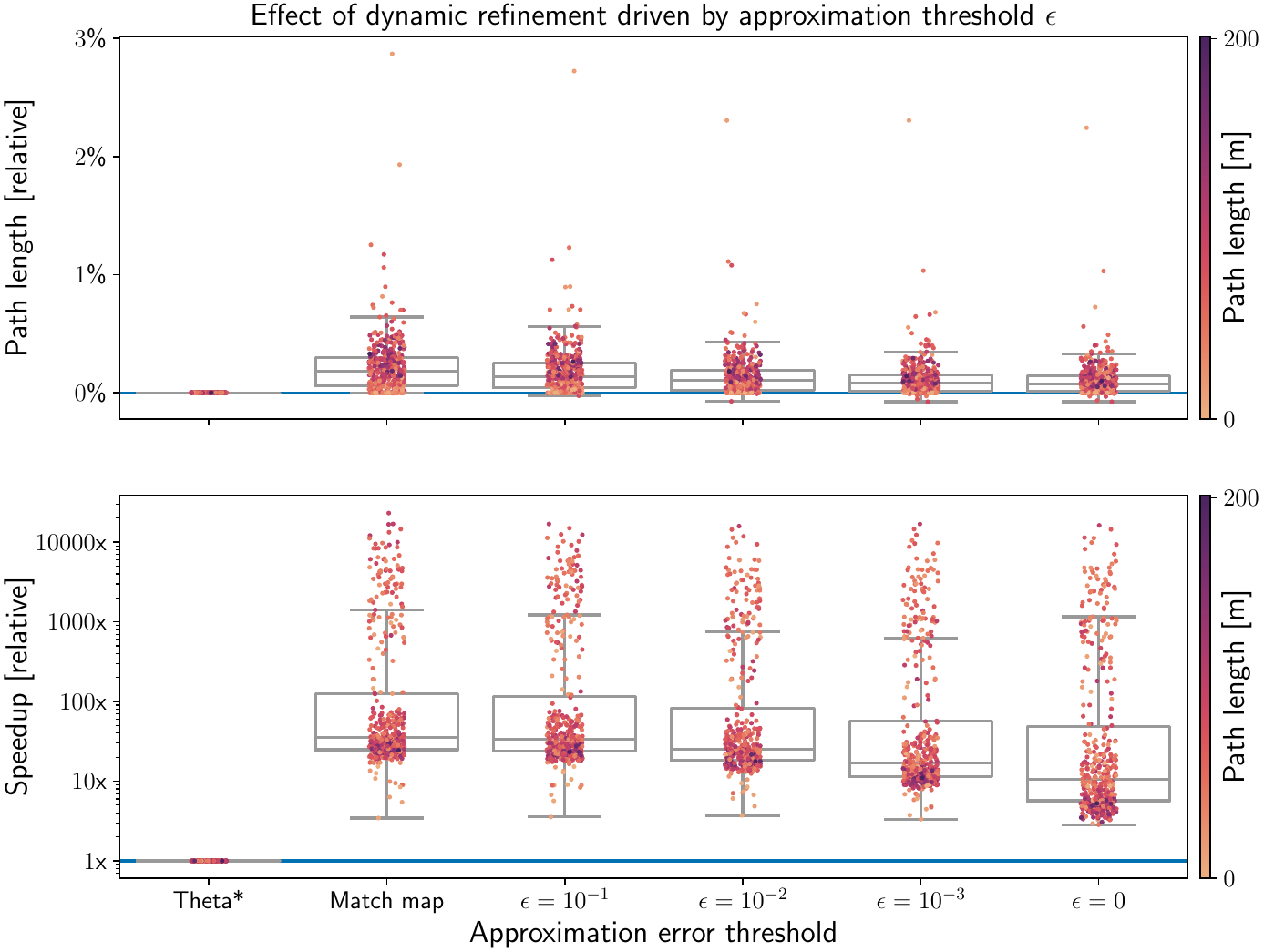}
    \caption{Ablation showing how the dynamic refinement strategy affects the path length and speedup (log scale) of our method relative to Theta* ({\color[HTML]{0073B2}blue} line). Reading the plot from left to right, we see that as the threshold is tightened, the path lengths decrease while runtime moderately increases.}
    \label{fig:global_planning/ablations_refinement}
\end{figure}

The results in \Cref{fig:global_planning/ablations_refinement} show that as the threshold $\epsilon$ is tightened, the path lengths gradually approach those of $\ThetaStar$. Similar to the initialization ablations, the error relative to $\ThetaStar$ is already low for $\MatchMap$ because very coarse free space leaves rarely occur in occupancy maps. This highlights that our multi-resolution, any-angle cost field formulation is inherently accurate, even at the moderate resolutions that dominate most free space. As $\epsilon$ is reduced, the relative path lengths generally remain within their $\epsilon$ suboptimality thresholds. However, some outliers persist, and the paths do not fully converge to $\ThetaStar$, even for $\epsilon=0$, due to the absence of initialization in this ablation, which limits the discovery of critical inflection points. In terms of runtime, reducing $\epsilon$ gradually increases computation, as smaller thresholds require more refinement steps.

\begin{table}
    \caption{Summary of ablations. Path lengths relative to Theta*.}
    \label{tab:global_planning/summary_ablations}
    \centering
    \begin{tabular}{llll}
        \multicolumn{1}{r}{} & \multicolumn{1}{r}{} & \multicolumn{2}{c}{Initialization} \\
        \multicolumn{2}{l}{Path length $\mu \pm \sigma$} & None & $\SI{10}{\centi\meter}$ \\
        \hline
        \multirow{2}{*}{Refinement} & None               & 0.23\% $\pm$ 0.43\% & 0.07\% $\pm$ 0.13\% \\
                                    & $\epsilon=10^{-2}$ & 0.16\% $\pm$ 0.32\% & \textbf{0.04}\% $\pm$ \textbf{0.12}\% \\
        \hline
    \end{tabular}
\end{table}

In conclusion, as shown in \Cref{tab:global_planning/summary_ablations}, initialization and refinement each reduce the mean and variance of path lengths, with their combination providing the greatest improvement.

\subsection{Comparisons with other planners}
The comparisons are performed on maps of four real environments ($\Mine$, $\Cloister$, $\Math$, and $\Park$ sequences of the Newer College Dataset~\cite{zhangMultiCameraLiDARInertial2022}) mapped with \textit{wavemap}~\cite{reijgwart2023wavemap} at \SI{10}{\centi\meter} resolution. These sequences represent constrained indoor ($\Mine$), mixed indoor-outdoor ($\Cloister$), large urban ($\Math$), and large vegetated environments ($\Park$). Obstacles were inflated by \SI{35}{\centi\meter} to account for the robot's radius. For each map, $100$ random collision-free start-goal pairs ($400$ total) were sampled. To assess how each planner handles unsolvable cases, infeasible queries were not filtered out.

We compare the success rates, path lengths, and execution times of our proposed multi-resolution planner to a representative set of search and sampling-based planners. In terms of search-based planners, we implemented fixed-resolution versions of $\AStar$~\cite{hart1968Astar}, $\ThetaStar$~\cite{daniel2010ThetaStar}, and $\LazyThetaStar$~\cite{nash2010LazyThetaStar}. For $\AStar$, we used the octile distance heuristic, which is consistent on 26-connected grids, as we found it to run up to $70\%$ faster than using the Euclidean distance heuristic. Additionally, we include the reference implementation
of $\OctreeLazyThetaStar$~\cite{faria2019efficient} as a multi-resolution baseline. For sampling-based planning, we used the RRTConnect~\cite{kuffner2000RRTConnect} and RRT*~\cite{karaman2011RRTStar} implementations from the Open Motion Planning Library~\cite{sucan2012ompl}. While $\RRTConnect$ terminates immediately once a path is found, RRT* does not. Therefore, we include three RRT* variants with increasing time budgets, namely $\RRT{0.1}$, $\RRT{1}$ and $\RRT{10}$. Note that $\RRTConnect$ is also limited to a maximum time budget of \SI{10}{\second}, to keep it from running forever when a planning query is infeasible.

We evaluate three variants of our multi-resolution planner: $\Ours$, $\OursLazy$, and $\OursFast$. $\Ours$ implements the baseline algorithm as described in the method section. To improve runtime further, $\OursLazy$ and $\OursFast$ incorporate lazy visibility checking, which reduces computational overhead by deferring visibility evaluations until necessary. These variants modify the baseline algorithm as detailed in \Cref{appendix:ours_with_lazy_checking}, applying the principles of LazyTheta*~\cite{nash2010LazyThetaStar}.
The specific settings we use for our three planner variants are:
\begin{itemize}
    \item \makebox[\widthof{$\OursLazy$:\,}][l]{$\Ours$:} $\epsilon=10^{-2}, r^\text{init}=\SI{10}{\centi\meter}$
    \item \makebox[\widthof{$\OursLazy$:\,}][l]{$\OursLazy$:} $\epsilon=10^{-2}, r^\text{init}=\SI{10}{\centi\meter}$, lazy visibility checks
    \item \makebox[\widthof{$\OursLazy$:\,}][l]{$\OursFast$:} $\epsilon=10^{-2}, r^\text{init}=\SI{40}{\centi\meter}$, lazy visibility checks
\end{itemize}

To ensure fair comparisons, all planners, including ours, use optimized data structures and subroutines. The fixed-resolution search-based planners store their cost fields using a hashed voxel block data structure~\cite{niessner2013real}, while our multi-resolution planner employs a hashed octree data structure~\cite{musethOpenVDBOpensourceData2013}. These planners and all RRT variants use \textit{wavemap}'s hierarchical occupancy map and multi-resolution ray tracer for fast traversability and visibility checking. As motivated in \Cref{sec:problem_statement}, we inflate all obstacles by the robot's radius, allowing it to be treated as a point. We configure $\OctreeLazyThetaStar$, which uses Octomap~\cite{hornungOctoMapEfficientProbabilistic2013} and a custom visibility checker, to match this setup. All experiments are run single-threaded on the same benchmarking server with an Intel i9-9900K CPU and \SI{64}{\giga\byte} of RAM.

\subsubsection{Success rates}
Starting with the success rates shown in \Cref{tab:global_planning/comparisons_success_rates}, all search-based planners perform equally well. As the start and goal pose pairs can contain infeasible planning queries, even complete planners may fail in some environments. For example, in $\Mine$ where none of the planners succeed in more than $88$ out of $100$ queries due to limited connectivity between areas.

Among the sampling-based planners, $\RRTConnect$ achieves the highest success rate, performing almost as well as the search-based planners. Its bidirectional tree growth and lack of rewiring provide an efficiency advantage over $\RRT{10}$, which comes in a close second (both planners operate within a maximum time budget of \SI{10}{\second}). $\RRT{1}$ trails slightly behind $\RRT{10}$ in simpler environments but struggles in maps that are large ($\Park$) or have narrow passages ($\Cloister$), as these scenarios require extensive sampling to ensure adequate coverage or density. Finally, $\RRT{0.1}$ performs reasonably well only in the $\Math$ environment, which consists of wide open spaces with good visibility.

\begin{table}
    \caption{Planning success rates per map for 100 randomly sampled queries each, including infeasible cases.}
    \label{tab:global_planning/comparisons_success_rates}
    \centering
    \begin{tabular}{lrrrr}
        Success rate (\%)    & Mine  & Cloister  & Math  & Park  \\
        \hline
        A*                 & \textbf{88}    & \textbf{100}   & \textbf{98}    & \textbf{99} \\
        Theta*             & \textbf{88}    & \textbf{100}   & \textbf{98}    & \textbf{99} \\
        LazyTheta*         & \textbf{88}    & \textbf{100}   & \textbf{98}    & \textbf{99} \\
        OctreeLazyTheta*   & \textbf{88}    & \textbf{100}   & \textbf{98}    & \textbf{99} \\
        RRTConnect         & \textbf{88}    & \underline{97} & \textbf{98}    & \underline{97} \\
        RRT* 0.1s          & 80             & 37             & 96             & 56 \\
        RRT* 1s            & \underline{85} & 52             & \underline{97} & 90 \\
        RRT* 10s           & \textbf{88}    & 87             & \textbf{98}    & 96 \\
        \textit{Ours}      & \textbf{88}    & \textbf{100}   & \textbf{98}    & \textbf{99} \\
        \textit{Ours Lazy} & \textbf{88}    & \textbf{100}   & \textbf{98}    & \textbf{99} \\
        \textit{Ours Fast} & \textbf{88}    & \textbf{100}   & \textbf{98}    & \textbf{99} \\
        \hline
    \end{tabular}
\end{table}

We verified that for every query where at least one planner succeeded, all search-based planners also succeeded. This empirical finding suggests that our multi-resolution planners maintain the completeness guarantee of their fixed-resolution counterparts. Additionally, there were no cases where a sampling-based planner found a solution that the search-based planners could not. This supports the idea that the adjacency graph of an occupancy map provides a reliable approximation of the solution space.

\subsubsection{Path length}
\begin{figure}[bt]
    \centering
    \includegraphics[width=\linewidth]{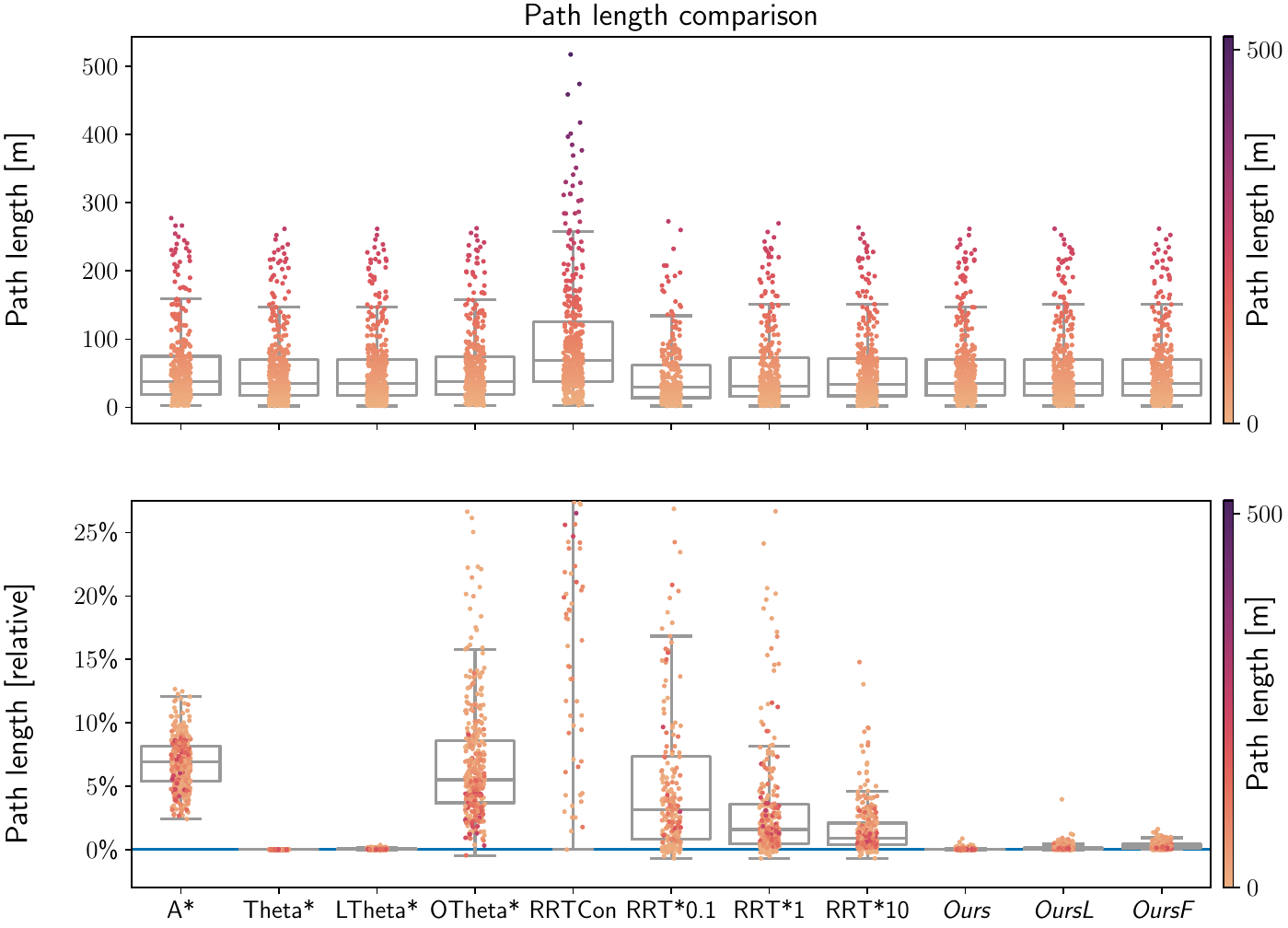}
    \caption{Path lengths for selected search- and sampling-based planners and three variants of our multi-resolution planner. The upper plot shows absolute lengths, and the lower plot shows lengths relative to $\ThetaStar$ ({\color[HTML]{0073B2}blue} line). Outliers for sampling-based planners are partially omitted in the lower plot. In particular, only the bottom few quantiles of $\RRTConnect$ are visible.}
    \label{fig:global_planning/comparisons_path_length}
\end{figure}

\begin{table}
    \caption{Average path lengths per map for queries where all planners succeeded.}
    \label{tab:global_planning/comparisons_average_path_lengths}
    \centering
    \begin{tabular}{lrrrr}
        Mean path length (m)    & Mine  & Cloister  & Math  & Park  \\
        \hline
        A*                 & 15.96             & 20.49             & 45.10             & 106.05  \\
        Theta*             & \textbf{14.87}    & \textbf{19.06}    & \textbf{42.20}    & \textbf{99.12} \\
        LazyTheta*         & \underline{14.88} & \textbf{19.06}    & 42.22             & \underline{99.14}  \\
        OctreeLazyTheta*   & 16.34             & 20.42             & 44.67             & 103.63  \\
        RRTConnect         & 30.49             & 39.49             & 73.83             & 155.89  \\
        RRT* 0.1s          & 17.50             & 19.59             & 44.05             & 106.52  \\
        RRT* 1s            & 15.81             & 19.35             & 43.04             & 101.41  \\
        RRT* 10s           & 15.16             & 19.19             & 42.61             & 100.04  \\
        \textit{Ours}      & 14.89             & \textbf{19.06}    & \underline{42.21} & 99.15  \\
        \textit{Ours Lazy} & 14.91             & \underline{19.07} & 42.26             & 99.21  \\
        \textit{Ours Fast} & 14.95             & 19.08             & 42.28             & 99.25  \\
        \hline
    \end{tabular}
\end{table}

Moving on to the path quality evaluations, we compare the average path lengths for all planners. To ensure fairness, only queries where all planners succeeded are included, avoiding bias toward planners that fail more often on longer paths. As shown in \Cref{tab:global_planning/comparisons_average_path_lengths}, $\ThetaStar$ consistently finds the shortest paths on all maps. $\LazyThetaStar$ follows closely, with paths only $0.03\%$ longer on average, and $\Ours$ achieves similar results. $\OursLazy$ and $\OursFast$ also perform well, with $\OursFast$ producing slightly longer paths but never exceeding $\ThetaStar$ by more than $0.5\%$. In contrast, the RRT* variants gradually increase in path length as their time budgets decrease. $\AStar$ and $\OctreeLazyThetaStar$ also yield noticeably longer paths, as $\AStar$ is constrained to a 26-connected grid and $\OctreeLazyThetaStar$ restricts paths to octree leaf centers. Finally, $\RRTConnect$ produces the longest paths, which, on average, are almost twice as long as those of $\ThetaStar$.

The absolute and relative path length distributions in \Cref{fig:global_planning/comparisons_path_length} provide additional insights. The evaluated path lengths ranged from $0$ to $500\si{\meter}$, with shorter paths being more frequent due to the smaller connected areas in maps like $\Mine$ and $\Cloister$. Most planners find reasonable paths, but $\RRTConnect$ stands out with significantly longer paths and high variance. $\RRT{0.1}$ appears to find slightly shorter paths on average, a bias explained by its failure to solve queries with distant start and goal pairs. 

For relative path lengths, $\ThetaStar$ consistently finds the shortest paths and is closely followed by $\LazyThetaStar$. While RRT* occasionally surpasses $\ThetaStar$, it is less consistent overall. For instance, $\RRT{10}$ features outliers with paths up to $1.8$ times longer than $\ThetaStar$. $\RRTConnect$ exhibits extreme variance, with paths up to $16.8$ times longer than $\ThetaStar$. This highlights the importance of RRT*’s tree rewiring for improving path quality. As predicted by Nash et al.~\cite{nash2010LazyThetaStar}, $\AStar$ paths are far from optimal, with most paths being at least $2\%$ longer and some reaching the theoretical worst-case of $13\%$. Similarly, $\OctreeLazyThetaStar$ can introduce significant detours, sometimes exceeding $25\%$. Finally, $\Ours$ closely matches $\ThetaStar$ on average, demonstrating high consistency and very few outliers. $\OursLazy$ introduces slight variability due to lazy visibility checking, producing paths that are marginally less direct. Also reducing the inflection point initialization resolution in $\OursFast$ results in slightly larger detours around obstacles. Nonetheless, paths produced by $\OursFast$ remain suitable for many practical applications.

\subsubsection{Runtime}
\begin{figure}[bt]
    \centering
    \includegraphics[width=\linewidth]{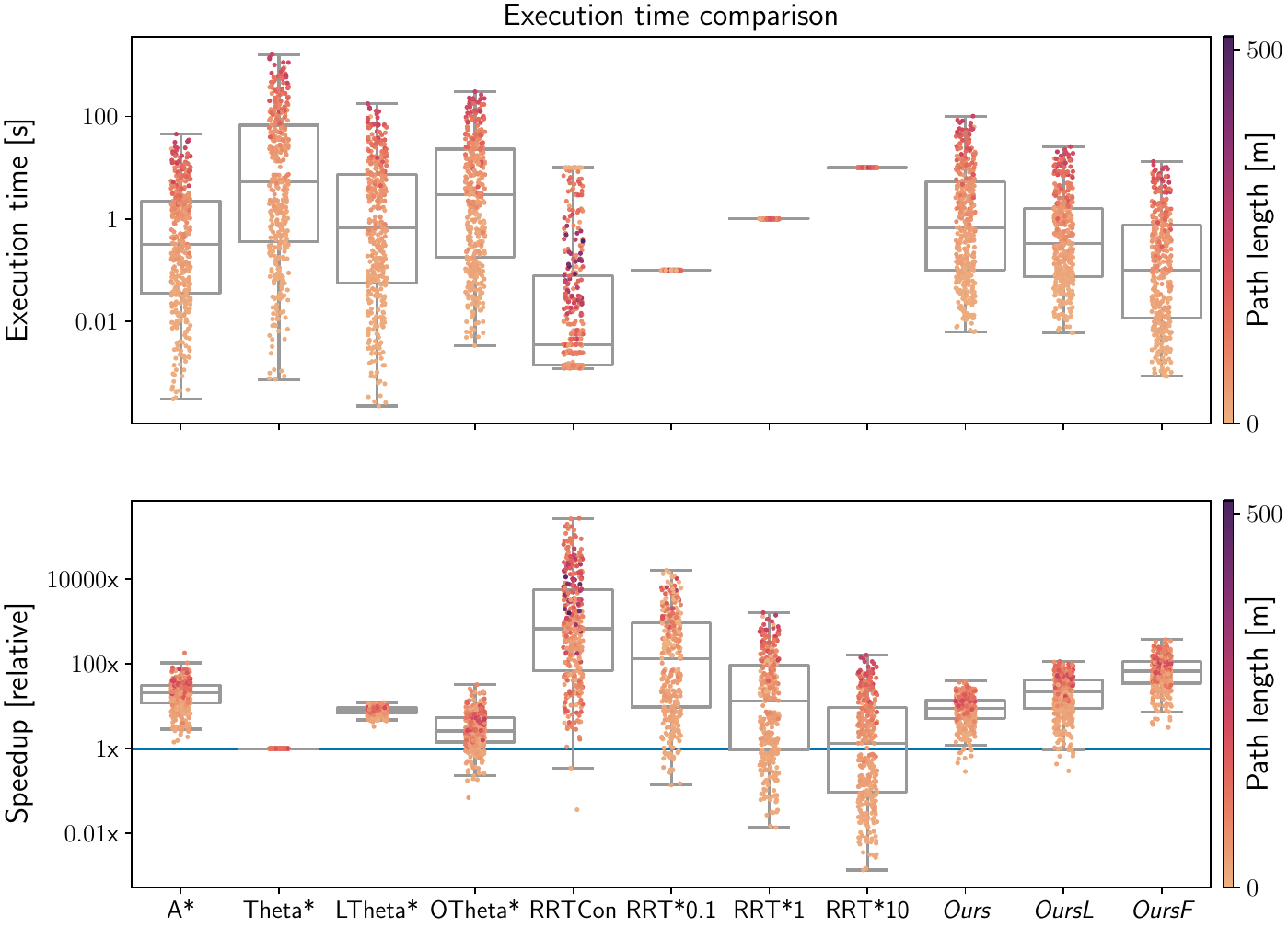}
    \caption{Execution times for selected search- and sampling-based planners and three variants of our multi-resolution planner. The upper plot shows absolute times (log scale), and the lower plot shows speedups relative to $\ThetaStar$ ({\color[HTML]{0073B2}blue} line).}
    \label{fig:global_planning/comparisons_runtime}
\end{figure}
\begin{table}
    \caption{Average execution times per map for 100 randomly sampled queries (RRT* variants omitted due to fixed runtimes).}
    \label{tab:global_planning/comparisons_average_runtimes}
    \centering
    \definecolor{gray}{gray}{0.6}
    \setlength\tabcolsep{3.8pt}
    \begin{tabular}{lrrrr}
        Mean execution time (s)&          Mine  &      Cloister  &          Math  &          Park  \\
        \hline
        A*                 & \underline{0.20} & 1.15             & 1.89             & 8.19  \\
        Theta*             & 2.70             & 30.29            & 61.18            & 281.38  \\
        LazyTheta*         & 0.42             & 3.66             & 6.89             & 31.51  \\
        OctreeLazyTheta*   & 2.88             & 10.79            & 13.12            & 64.94  \\
        RRTConnect         & 1.24             & 1.63             & \textbf{0.21}    & \textbf{0.39} \\
        \textit{Ours}      & 0.51             & 2.69             & 3.86             & 20.22  \\
        \textit{Ours Lazy} & \underline{0.20} & \underline{0.82} & 1.12             & 5.72  \\
        \textit{Ours Fast} & \textbf{0.10}    & \textbf{0.36}    & \underline{0.51} & \underline{2.76}  \\
        \hline
    \end{tabular}
\end{table}

The last metric we evaluate is execution time, starting with the average runtime of each planner in each environment, as shown in \Cref{tab:global_planning/comparisons_average_runtimes}. All runs are included in the averages to capture both successful and unsuccessful queries, while the planning times of the RRT* variants are not listed as they are constant. $\ThetaStar$ is the slowest planner by a large margin. Enabling lazy visibility checking ($\LazyThetaStar$) improves its runtime by $6$ to $9$ times, but it remains significantly slower than $\AStar$. Interestingly, $\OctreeLazyThetaStar$ is not faster than $\LazyThetaStar$, indicating that octree-based representations alone do not outperform optimized fixed-resolution methods. However, substantial speedups can be achieved through careful multi-resolution design. On average, $\Ours$ is $8$ times faster than $\ThetaStar$ in confined environments and $15$ times faster in large open spaces. $\OursLazy$ achieves similar gains, being $2$ to $6$ times faster than $\LazyThetaStar$. $\OursFast$ and $\RRTConnect$ are the fastest overall. $\OursFast$ is up to $12$ times faster than $\RRTConnect$ in confined environments like $\Mine$, while $\RRTConnect$ is up to $7$ times faster in large open spaces like $\Park$.

\Cref{fig:global_planning/comparisons_runtime} shows runtime distributions, absolute with logarithmic scale (top) and relative to $\ThetaStar$ (bottom). For search-based planners, execution time is strongly correlated with path length. In contrast, $\RRTConnect$ shows no meaningful correlation, and the RRT* variants maintain constant runtimes. Notably, the speedup of all planners over $\ThetaStar$ increases with path length, reflecting $\ThetaStar$’s poor scalability.

Looking at our proposed planners, $\Ours$ and $\OursLazy$ are rarely slower than $\ThetaStar$ and achieve speedups of up to $60$ and $100$ times, respectively. $\OursFast$ consistently outperforms all other search-based planners, with speedups ranging from $5$ to $800$ times. We can also see that our method's runtimes are more predictable than those of sampling-based methods, in part due to being able to detect infeasible queries. This result underscores the effectiveness of leveraging multi-resolution to balance efficiency and path quality.

%% file: sections/limitations.tex
\section{Limitations}

The main limitation of our method is its restriction to Euclidean cost formulations. This is because, like Theta*, our multi-resolution extension relies on the triangle inequality to find any-angle paths efficiently~\cite{nash2010LazyThetaStar}. We also focused on Euclidean workspaces, to reliably and efficiently provide global paths as inputs to trajectory optimizers or local planners in navigation tasks~\cite{nieuwenhuisen2016localMulti,zhou2021raptor,chen2022real}.


%% file: sections/conclusion.tex
\section{Conclusion}
In this paper, we presented \textit{wavestar}, a search-based global planning method for Euclidean workspaces that utilizes an octree-like structure to improve planning speed in occupancy maps. We extend the ideas from any-angle planners to hierarchical representations to exploit spatial sparsity by generalizing the concept of inflection points from fixed-resolution grids to a hierarchical representation.

Extensive evaluations and comparisons to search-based methods show that we achieve paths of competitive quality but at a substantially reduced computational cost. This demonstrates that exploiting the inherent sparsity of real environments does not significantly impact accuracy, while providing significant computational benefits. Additionally, in contrast to sampling-based methods, our approach can detect when a query is infeasible, while also producing high-quality paths. Overall, these results show that our approach combines the benefits and guidance of search-based methods with the speed of sampling-based methods. This makes it suitable as a first step for many navigation systems, where a coarse initial path through 3D space is needed.

%% file: sections/acknowledgements.tex
\section*{Acknowledgments}
We would like to thank Helen Oleynikova, Alexander Liniger and the reviewers for their valuable suggestions and feedback on the manuscript.


%% file: sections/appendix.tex
\subsection{A brief introduction to Theta*}
\label{appendix:theta_star}
Given that our method's cost field formulation and search algorithm can be seen as a multi-resolution extension of Theta*, we briefly summarize the algorithm and relevant terminology in this section. Theta*~\cite{daniel2010ThetaStar} is an any-angle path-finding extension to the A*~\cite{hart1968Astar} search algorithm. Just like A*, it only propagates information along grid edges. The key distinction between the two search algorithms lies in how they select each vertex's predecessor. Since A* only considers each vertex's direct neighbors, the paths it returns are strictly composed of grid edges. Theta* additionally considers connections to each direct neighbor's best predecessor if it is within the vertex's line of sight. This allows Theta* to deviate from the grid and find paths up to $\approx13\%$ shorter than those found by A*, at the cost of increased runtime due to the additional visibility checks.

\begin{algorithm}
    \SetAlgoLined
    \SetNoFillComment
    \DontPrintSemicolon

    \SetKwProg{Fn}{Function}{ is}{end}
    \SetKwFunction{UpdateVertex}{UpdateVertex}
    \SetKwFunction{UpdateCost}{UpdateCost}
    \SetKwFunction{CheckVertex}{CheckVertex}
    
    $\OpenQueue \gets \emptyset$\;
    $\ClosedSet \gets \emptyset$\;
    $g(s^\text{start}) \gets 0$\;
    $\Predecessor(s^\text{start}) \gets s^\text{start}$\;
    $\OpenQueue.insert(s^\text{start},g(s^\text{start}) + h(s^\text{start}))$\;
    \While{$\OpenQueue \neq \emptyset$}{%
        $s \gets \OpenQueue.pop()$\;
        \If{$s = s^\text{goal}$}{\label{algoline:global_planning/fixed_res_heuristic_guided_graph_search/expanded_goal}%
            \Return{\PathFound}
        }
        $\ClosedSet \gets \ClosedSet \cup \{s\}$\;
        \ForEach{$s' \in \text{neighbors}(s)$}{%
            \If{$s' \notin \ClosedSet$}{\label{algoline:global_planning/fixed_res_heuristic_guided_graph_search/skip_closed}%
                \If{$s' \notin \OpenQueue$}{%
                    $g(s') \gets \infty$\;
                    $\Predecessor(s') \gets \text{NULL}$\;
                }
                $\UpdateVertex(s, s')$\;
            }
        }
    }
    \Return{\NoPathFound}
    \BlankLine
    \Fn{$\UpdateVertex(s, s')$}{%
        Status $\gets \UpdateCost(s, s')$\;
        \If{Status = \Changed}{%
            \If{$s' \in \OpenQueue$}{%
                $\OpenQueue.remove(s')$\;
            }
            $\OpenQueue.insert(s', g(s') + h(s'))$\;
        }
    }
    
    \caption{Heuristic-guided search over vertices}
    \label{algo:global_planning/fixed_res_heuristic_guided_graph_search}
\end{algorithm}

The main loop of A* and Theta*, shown in \Cref{algo:global_planning/fixed_res_heuristic_guided_graph_search}, is identical. Both search algorithms store two values per vertex, namely the vertex's cost-to-come ($g$ cost) and an index or pointer to its best $\Predecessor$. Furthermore, both algorithms use a priority queue ($\OpenQueue$) to expand vertices in order of their minimum $f$ score, where $f(s) = g(s) + h(s)$ with $h(s)$ a consistent heuristic function. Using a consistent heuristic guarantees that a node is only expanded from the queue once its optimal $g$ cost and $\Predecessor$ have been found~\cite{russell2010ArtificialIntelligenceModern}. A closed set ($\ClosedSet$) can therefore be used to track and explicitly skip updates of already expanded nodes (\Cref{algoline:global_planning/fixed_res_heuristic_guided_graph_search/skip_closed}). It also means that both algorithms can terminate immediately once the goal vertex $s^\text{goal}$ is expanded (\Cref{algoline:global_planning/fixed_res_heuristic_guided_graph_search/expanded_goal}). Since the paths found by A* can only contain edges of the 26-connected grid, using the octile distance to the goal, $h(s)=\lVert s^\text{goal}-s \rVert_\text{oct}$, is consistent. In contrast, Theta* must use the Euclidean distance, $h(s)=\lVert s^\text{goal}-s \rVert_{2}$, because its paths are not constrained to the grid's edges.

\begin{algorithm}
    \SetAlgoLined
    \SetNoFillComment
    \DontPrintSemicolon

    \SetKwProg{Fn}{Function}{ is}{end}
    \SetKwFunction{UpdateCost}{UpdateCost}
    \SetKwFunction{CheckVertex}{CheckVertex}

    \Fn{$\UpdateCost(s, s')$}{%
        \If{$g(s) + c(s, s') < g(s')$}{%
            $\Predecessor(s') \gets s$\;
            $g(s') \gets g(s) + c(s, s')$\;
            \Return{\Changed}\;
        }
        \Return{\Unchanged}\;
    }
    
    \caption{Definitions for A*}
    \label{algo:global_planning/fixed_res_a_star}
\end{algorithm}

\begin{algorithm}
    \SetAlgoLined
    \SetNoFillComment
    \DontPrintSemicolon

    \SetKwProg{Fn}{Function}{ is}{end}
    \SetKwFunction{UpdateCost}{UpdateCost}
    \SetKwFunction{CheckVertex}{CheckVertex}
    \SetKwFunction{LineOfSight}{LineOfSight}

    \Fn{$\UpdateCost(s, s')$}{%
        $s^p \gets \Predecessor(s)$\;
        \eIf{$\LineOfSight(s^p, s')$}{%
            \tcp{Ray traced connection}
            \If{$g(s^p) + c(s^p, s') < g(s')$}{%
                $\Predecessor(s') \gets s^p$\;
                $g(s') \gets g(s^p) + c(s, s')$\;
                \Return{\Changed}\;
            }
        }{%
            \tcp{Direct neighbor connection}
            \If{$g(s) + c(s, s') < g(s')$}{%
                $\Predecessor(s') \gets s$\;
                $g(s') \gets g(s) + c(s, s')$\;
                \Return{\Changed}\;
            }
        }
        \Return{\Unchanged}
    }
    
    \caption{Definitions for Theta*}
    \label{algo:global_planning/fixed_res_theta_star}
\end{algorithm}

As highlighted earlier, the key difference between A* and Theta* is how they find each vertex's best predecessor. When expanding vertex $s$, the function $\UpdateCost$ is called for each neighboring vertex $s'$ to check if using $s$ could lead to a shorter path. In that check A* only considers connecting $s'$ directly to $s$ (\Cref{algo:global_planning/fixed_res_a_star}, note that $c(s^a, s^b)$ refers to the edge cost between two vertices $s^a$ and $s^b$). As shown in \Cref{algo:global_planning/fixed_res_theta_star}, Theta* considers connections from $s'$ to both $s$ and $\Predecessor(s)$. By virtue of the triangle inequality, a connection to $\Predecessor(s)$ -- when available -- is guaranteed to yield a candidate $g$ cost that is equal to or lower than a direct connection to $s$. Therefore, Theta* only considers connecting to direct neighbor $s$ when its $\Predecessor(s)$ is not visible from $s'$.

\subsection{Lazy visibility checking extension}
\label{appendix:ours_with_lazy_checking}
As demonstrated by LazyTheta*\cite{nash2010LazyThetaStar}, lazy visibility checking can improve the runtime of Theta*~\cite{daniel2010ThetaStar} by over an order of magnitude without significantly increasing path length. This technique can also be applied to our multi-resolution any-angle planner to achieve similar benefits. To implement this extension, we modify the $\UpdateCost$ function (\cref{algo:global_planning/multi_res_theta_star}) by assuming that $\LineOfSight(s^p, \mathcal{V}')$ is always true during the initial cost update (\Cref{algoline:global_planning/multi_res_theta_star/has_line_of_sight}). When the search later expands $\mathcal{V}'$ (\cref{algo:global_planning/multi_res_heuristic_guided_graph_search} \cref{algoline:global_planning/multi_res_heuristic_guided_graph_search/expand_v}), this assumption is verified. If the visibility check fails, the $\Predecessor$ of $\mathcal{V}'$ is updated to the best direct neighbor, introducing only a minor detour as the best neighbor's center is typically very close to the optimal path. The remainder of the algorithm remains unchanged.

\subsection{Formal statements}
\label{appendix:formal_statements}
To the authors' best knowledge, no proofs or formal guarantees on completeness, optimality, or time complexity have appeared in prior literature for either Theta*~\cite{daniel2010ThetaStar} or LazyTheta*~\cite{nash2010LazyThetaStar}. In the statements that follow, we therefore discuss observations applying to both Theta* and our multi-resolution extension.

\subsubsection{Completeness}
Theta* operates on a supergraph of the grid's 26-connected adjacency graph considered by A*, as its additional line-of-sight check can only \textit{add} edges. Since each vertex receives at most one additional edge, the supergraph's branching factor remains finite. Furthermore, all edge weights are non-zero. Therefore, Theta* preserves A*'s resolution completeness guarantee.

Our method losslessly converts the traversability grid into octree leaves and defines any two leaves as adjacent if they are both traversable and touch in any form, i.e., have an overlapping face, edge, or corner. For any two connected vertices on the fixed-resolution grid, the respective octree leaves that cover them will also be connected. Our representation, therefore, maintains the connectivity of the original space, preserving the resolution completeness of running A* on the original high-resolution grid.

\subsubsection{Optimality}
We build on the previous section's insight that Theta* operates on a supergraph of A*'s graph. New edges are only introduced if they reduce the evaluated vertex's cost-to-come. Moreover, the chosen Euclidean distance heuristic remains admissible and consistent. Theta* is therefore guaranteed to find paths that are at least as good as those of A* and, as proven in~\cite{nash2010LazyThetaStar}, it can find solutions that are up to $\approx13\%$ shorter.

Naively applying A* or Theta* to the center points of an octree's leaves results in paths that can be arbitrarily suboptimal with respect to their fixed-resolution counterparts, as illustrated in \Cref{fig:global_planning/media/octree_a_star_wrong_homotopy}. Our method addresses this problem by leveraging an octree while still considering the high-resolution vertices covered by each leaf.
Our algorithm's initialization procedure (\Cref{subsec:global_planning/initializing_inflection_points}) guarantees that all potentially optimal predecessors, at the chosen initialization resolution, are considered. During the multi-resolution search, our dynamic refinement procedure (\cref{subsec:global_planning/dynamic_refinement}) increases the resolution of each subvolume until the cost to reach each high-resolution vertex it covers is suboptimal by a factor of at most $2\epsilon$. This factor of two follows from the fact that \Cref{eq:global_planning/cost_field_approximation_error_invariant} is applied pairwise between all successive candidates. Since the goal is itself a vertex, the total path length will be within $2\epsilon$ of Theta* and, by extension, A*.

Note that setting $\epsilon=0$ and matching the initialization resolution to the fixed-resolution grid yields paths whose length is almost identical to those of Theta*, but that sometimes differ in terms of chosen predecessors. We believe this follows from the fact that Theta* is not guaranteed to find the optimal \textit{any-angle} Euclidean shortest path. The predecessors chosen by Theta* and our algorithm are sensitive to the order in which the vertices or subvolumes are expanded, and this order will differ slightly due to small numerical and discretization differences in our multi-resolution formulation.


\subsubsection{Time Complexity}
Our method leverages a data structure that combines hash maps with fixed-maximum-depth octrees~\cite{musethOpenVDBOpensourceData2013}, achieving the same $O(1)$ access and insertion complexity as regular grids. Consequently, the time complexity of expanding a subvolume in our method matches Theta*'s vertex expansion complexity. Theta* performs up to 26 predecessor-visibility checks per expanded vertex, each potentially having a complexity that grows linearly in the checked distance. However, very long-distance visibility checks can be avoided without significantly affecting path length, since the detour introduced by splitting a long edge into two and aligning their middle vertex to the grid is negligible. Our implementations of Theta* and our multi-resolution extension therefore reject checks beyond a fixed maximum distance, ensuring their worst-case runtime complexity matches that of A* on a regular grid. Experimentally, we demonstrate that our method's runtime is substantially lower in practice (\cref{tab:global_planning/comparisons_average_path_lengths}).
